\documentclass[runningheads]{llncs}
\usepackage{graphicx}

\usepackage{tikz}
\usepackage{comment}
\usepackage{amsmath,amssymb} 
\usepackage{color}

\usepackage[accsupp]{axessibility}  

\usepackage{threeparttable}
\usepackage{adjustbox}

\usepackage{eqparbox}
\usepackage{arydshln}
\usepackage{wrapfig}

\DeclareMathOperator{\EX}{\mathbb{E}}

\newcommand{\distas}[1]{\mathbin{\overset{#1}{\kern\z@\sim}}}%
\newsavebox{\mybox}\newsavebox{\mysim}
\newcommand{\distras}[1]{%
	\savebox{\mybox}{\hbox{\kern3pt$\scriptstyle#1$\kern3pt}}%
	\savebox{\mysim}{\hbox{$\sim$}}%
	\mathbin{\overset{#1}{\kern\z@\resizebox{\wd\mybox}{\ht\mysim}{$\sim$}}}%
}

\begin{document}
\pagestyle{headings}
\mainmatter
\def\ECCVSubNumber{6084}  


\title{Domain Adaptation in LiDAR Semantic Segmentation via Alternating Skip Connections and Hybrid Learning} 


\titlerunning{Domain Adaptation in LiDAR Semantic Segmentation}
%

\author{Eduardo R. Corral-Soto\and
Mrigank Rochan\and
Yannis Y. He\and
Shubhra Aich\and
Yang Liu\and
Liu Bingbing  }


%
\authorrunning{Eduardo R. Corral-Soto et al.}
%

\institute{Huawei Noah's Ark Lab, Canada\\
\email{\{eduardo.corral.soto, mrigank.rochan1, yannis.yiming.he, shubhra.aich1, yang.liu9, liu.bingbing\}@huawei.com}}



\maketitle

\begin{abstract}
In this paper we address the challenging problem of domain adaptation in LiDAR semantic segmentation. We consider the setting where we have a fully-labeled data set from source domain and a target domain with a few labeled and many unlabeled examples. We propose a domain adaption framework that mitigates the issue of domain shift and produces appealing performance on the target domain. To this end, we develop a GAN-based image-to-image translation engine that has generators with alternating connections, and couple it with a state-of-the-art LiDAR semantic segmentation network. Our framework is hybrid in nature in the sense that our model learning is composed of self-supervision, semi-supervision and unsupervised learning. Extensive experiments on benchmark LiDAR semantic segmentation data sets demonstrate that our method achieves superior performance in comparison to strong baselines and prior arts. 

\keywords{Domain adaptation, LiDAR, Perception, Autonomous Driving}
\end{abstract}

\section{Introduction and Prior Work} \label{Sec:Intro_Related_work}
Recently, deep neural networks for LiDAR perception in autonomous driving and robotics have shown remarkable promise. The key to this success lies in the availability of large, manually labeled training datasets. Though these networks achieve promising performance on unseen data from the training domain, they fail to perform equally well on data from a different domain, mainly due to domain shift. For instance, a model trained on point cloud data captured from a LiDAR sensor yields very inferior performance on point cloud data obtained from a different LiDAR sensor. A solution for this is to collect a labeled training data set from the new domain and then utilize this dataset to re-train the network. To overcome this problem and mitigate the impact of domain shift, we propose a domain adaptation framework. We consider that we are given two data sets captured by different LiDAR sensors that capture the environment differently. The first one (source domain) is fully-labeled, while the second one (target domain) has only a small number of labeled frames and the rest unlabeled. Our goal is to train a LiDAR semantic segmentation network that generalizes well and achieves good performance on unseen data from the target domain. There exist a number of techniques in the literature that aim to solve this problem. For instance, transfer learning and fine-tuning methods ~\cite{bengio2012deep}, ~\cite{goodfellow2016deep}, ~\cite{pan2009survey}, ~\cite{tajbakhsh2016convolutional}, ~\cite{corral2020understanding}, ~\cite{olivas2009handbook}, ~\cite{caruana1994learning}, are commonly applied techniques where the network is first trained on the source data set, and then further trained (i.e. fine-tuned) using the available labeled target frames. In addition, there are domain adaptation based approaches that focus on reducing the domain shift between two or more domains. The taxonomy papers ~\cite{wang2018deep} and ~\cite{triess2021survey} extensively discuss various domain adaptation methods and classify them into: 1) Domain-invariant data representation methods ~\cite{rist2019cross}, ~\cite{alonso2020domain}, mainly based on hand-crafted data preprocessing to move different domains into a common representation (e.g., LiDAR data rotation and normalization), 2) Domain-invariant feature learning for finding a common representation space for the source and target domains ~\cite{jiang2020lidarnet}, ~\cite{jaritz2020xmuda}, 3) Discrepancy-based normalization statistics methods that attempt to align the domain distributions by a normalization of the mean and variance of activations. These include methods like Maximum Mean Discrepancy (MMD) ~\cite{gretton2006kernel} and DeepCORAL ~\cite{sun2016deep}, which minimize the global mean or covariance matrix discrepancy between the source and target domains, 4) Reconstruction-based methods, which use auxiliary reconstruction tasks to encourage feature invariance, and 5) Domain mapping adversarial methods, where source data is transformed, tipically using GANs ~\cite{goodfellow2014generative} to appear like target data ~\cite{saleh2019domain}, ~\cite{sallab2019lidar}, ~\cite{zhao2021epointda}. Some methods leverage simulated synthetic frames ~\cite{sallab2019lidar}, ~\cite{sallab2019unsupervised}, where a CycleGAN ~\cite{zhu2017unpaired} is trained using unpaired simulated and real projected bird's eye view (BEV) images to generate pseudo-labeled simulated data for off-line training (not end-to-end) of a BEV YOLOv3 ~\cite{redmon2018yolov3} object detection network. Fore example, SqueezeSegV2 \cite{wu2019squeezesegv2} uses simulators to generate large quantities of labeled spherical projections of synthetic LiDAR data to train perception models. Another method in ~\cite{langer2020domain} uses SLAM to fuse sequential LiDAR scans from the source data set into a dense mesh to sample and render semi-synthetic scans (to simulate data from a sensor with a lower number of channels) that match those from the target data set, improving the segmentation performance without the need of labeling. 
Other methods such as LiDARNet ~\cite{jiang2020lidarnet} and LCP ~\cite{corral-soto2021lcp} integrate CycleGAN into a task-specific network for training on real LiDAR source and target data sets. LiDARNet operates on LiDAR range images (spherical projection) concatenated with reflectivity, 2D coordinates, and normal maps to train a segmentation network.  LCP integrates and adapts CycleGAN to operate with  $64$-channel feature pseudo-images for the PointPillars 3D object detector ~\cite{lang2019pointpillars}.
In this paper, we present a novel image-to-image translation engine which is inspired by CycleGAN, but has no cycle connections and cycle consistency losses. We couple our translation engine with a state-of-the-art LiDAR semantic segmentation network ~\cite{cortinhal2020salsanext}, resulting in an integrated domain adaptation architecture we call HYLDA that can be trained end-to-end. 
To train the semantic segmentation network we adopt a diverse set of learning paradigms, including: 1) self-supervision on a simple auxiliary reconstruction task using all available raw (unlabeled) frames from the source and target domains, 2) semi-supervised training using the few available labeled target domain frames, and, 3)  unsupervised training on the fake translated images generated by the image-to-image translation engine using labels from the source domain, and no labels from the target domain.

The following is a summary of our contributions: \emph{1) An image-to-image translation engine} with generators consisting of separate encoder and decoder with alternating skip connections that allows us to make use of different learning paradigms, including self-supervision and adversarial, in its training, \emph{2) A hybrid learning LiDAR perception task stage} composed of encoder-decoder semantic segmentation networks, where the network is incrementally trained through self-supervision from an auxiliary reconstruction task, semi-supervision using a small number of labeled frames available, and lastly, in unsupervised manner from fake translated images from the image-to-image translation engine, \emph{3) We present a thorough evaluation section.} where we compare HYLDA against strong baselines and prior arts on three publicly-available LiDAR semantic segmentation data sets. We demonstrate that the end result from our proposed domain adaptation framework improves generalization of the semantic segmentation model on the unseen validation set from the target domain. We also present ablation study and an analysis highlighting the impact of our domain adaptation on the mean and covariance matrix from the translated data.


\section{Problem Being Addressed} \label{Sec:problem_definition}
\begin{figure*} 
	\centering	
	\includegraphics[width=0.5\columnwidth, trim={0cm 11cm 2cm 0cm},clip]{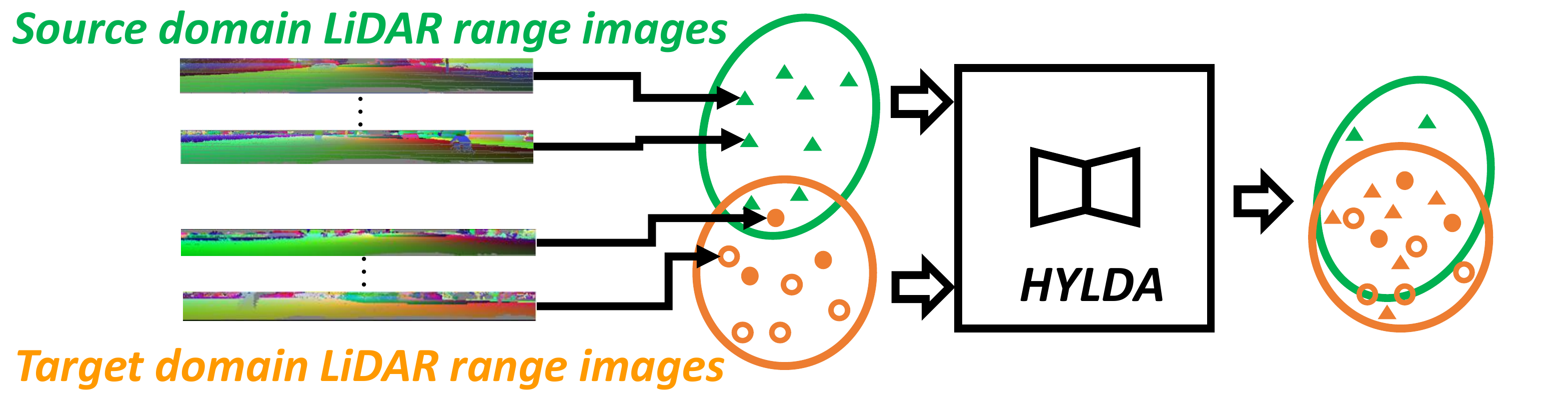} 		
 \caption{High-level illustration of our domain adaptation framework. The source domain is fully labeled, whereas the target domain has a small number of labeled examples (filled circles) and many unlabed examples (empty circles) }
\label{hylda_high_level}
\end{figure*}

Due to the domain shift induced by the sensor and geographical differences in the given data sets, a model trained on the source data set does not usually generalize well when evaluated directly on the target data set, as shown by rows $2$ and $15$ from Table \ref{table:kitti_vs_nuscenes_eval}. Therefore, we formulate our problem as follows: Given two data sets captured by different LiDAR sensors at different geographic locations, where the first one is fully-labeled, and the second one has only a small number of labeled frames and the rest unlabeled, we seek to design an architecture and training strategy that result in a trained semantic segmentation model with improved generalization on validation data from the target domain. Fig. \ref{hylda_high_level} provides a high-level illustration of our proposed framework.

\section{Our Method}
\begin{figure*} 
	\centering	
		\includegraphics[width=1.0\columnwidth]{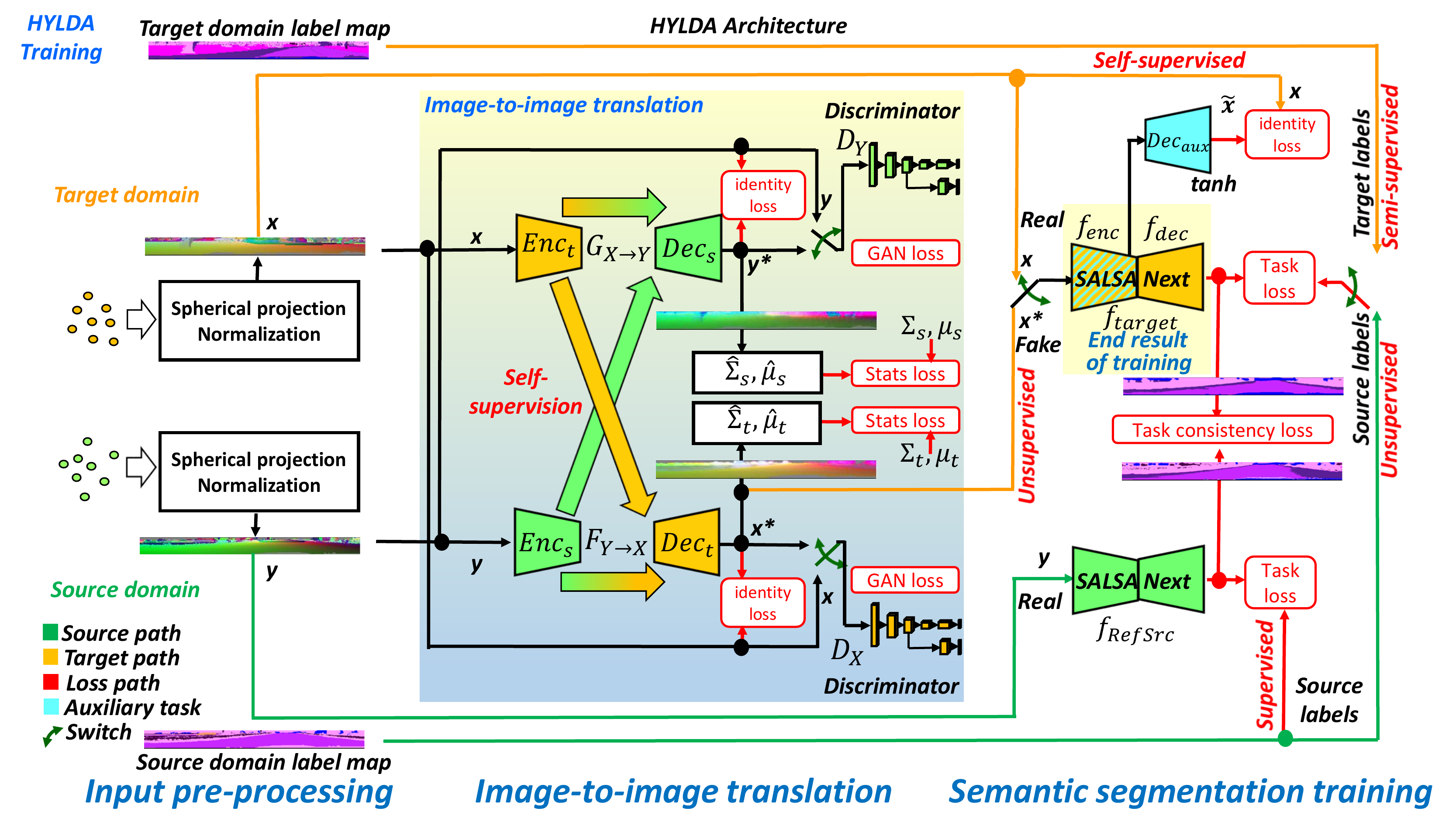} 		
	\caption{HYLDA integrated end-to-end architecture, which has three basic stages: 1) Input pre-processing, 2) Image-to-image translation engine, and 3) Task stage composed of LiDAR semantic segmentation networks. A step-by-step explanation of the HYLDA training steps can be found in the supplementary materials 		     				
	\label{HYLDA_architecture} }
\end{figure*}

Our proposed HYLDA architecture, shown in Fig. \ref{HYLDA_architecture}, has three basic stages: 1) Input pre-processing, 2) Image-to-image translation engine, and 3) Task stage composed of LiDAR semantic segmentation networks. 

\subsection{Input Pre-Processing} \label{Sec:preprocessing}
The inputs to HYLDA are two (source and target) LiDAR 3D point clouds and available point-level semantic labels. The point clouds are converted and normalized into spherical projection range view (RV) images of size $64$ $\times$ $2048$ $\times$ $5$ (X, Y, Z, range, remission) as done in ~\cite{cortinhal2020salsanext}, ~\cite{wu2019squeezesegv2}, ~\cite{gerdzhev2020tornado}, ~\cite{rochan2021unsupervised} to enable the use of 2D convolutional layers.

\subsection{Image-to-Image Translation Engine} \label{Sec:image_to_image_engine}
Our image-to-image translation engine is inspired by the CycleGAN method ~\cite{zhu2017unpaired}, where mapping functions (generators),  $G \colon X \to Y$, and $F \colon Y \to X$ are learned using two discriminators $D_X$ and $D_Y$ and adversarial training. Here $Y$ is the source domain, and $X$ is the target domain as shown in Fig. \ref{HYLDA_architecture}. 

\emph{Split generator encoder and decoder with alternating skip connections.} As discussed in ~\cite{zhu2017unpaired}, a generator trained with adversarial training can produce degenerate outputs by mapping a set of input source domain images onto random permutations of target domain images. To address this problem, CycleGAN introduced cycle consistency losses to translate the outputs from the generators back into their original domain to regularize the mappings $G$ and $F$. In contrast with CycleGAN, our proposed image-to-image translation engine \emph{does not} employ cycle connections nor cycle consistency losses. Instead, we propose to regularize the mappings $G$ and $F$ via self-supervision as explained next. With this goal in mind, we split each generator ($F_{Y \to X}$ and $G_{X \to Y}$ in Fig. \ref{HYLDA_architecture}) into separate independent encoder and decoder networks, and introduce \emph{alternating} skip connections. This design allows us to make use of different learning paradigms to train our generator encoders and decoders, including a) self-supervision where we \emph{cross}-connect the skip connections between the encoder and decoder from different generators, and b) adversarial GAN training, using \emph{intra}-generator skip connections between encoder and decoder of the same generator.

\emph{Discriminator}.
To classify image patches as either real or fake, we use the same CycleGAN Patch-GAN discriminators for adversarial training, however, with a minor modification. Inspired by the TSIT method for camera images ~\cite{jiang2020tsit}, we modify the discriminator to introduce one additional classification head connected to an earlier feature layer from the \emph{same} Patch-GAN network to ``criticize", at two resolutions, the fake outputs from the generator to provide more detailed feedback via back-propagation.


\emph{Domain statistics loss}.
To encourage the generators $F_{Y \to X}$ and $G_{X \to Y}$ to output fake translated images that adhere to the underlying statistics from the \emph{real} input domain they are translated into, we implement a statistics loss that combines DeepCORAL ~\cite{sun2016deep} and MMD ~\cite{gretton2006kernel}, by taking into account both the covariance matrix and mean. We start by precomputing the covariance matrix and mean image ($\Sigma_{s}$, $\mu_{s}$) for the \emph{whole} source training data set, and available target domain frames ($\Sigma_{t}$,$\mu_{t}$). During training, the mean and covariance matrix ($\hat{\Sigma}_{s}$, $\hat{\mu}_{s}$), and ($\hat{\Sigma}_{t}$,$\hat{\mu}_{t}$) are computed from each batch of fake translated images, and compared against their correponding pre-computed targets (as shown in Fig. \ref{HYLDA_architecture}) using an $L_1$ norm loss as described in section \ref{Sec:training_and_losses}. 

\subsection{Task Stage: Semantic Segmentation } \label{Sec:task_stage}
We focus on the LiDAR semantic segmentation task, and select the well-known SalsaNext model ~\cite{cortinhal2020salsanext} to develop our domain adaptation method. SalsaNext receives a LiDAR 3D point cloud as an input, and converts it into a $64$ $\times$ $2048$ $\times$ $5$ range view image as described in Section \ref{Sec:preprocessing}. Its output is a $64$ $\times$ $2048$ predicted pixel-level semantic class label map. We modified SalsaNext to support the $11$ semantic classes described in section \ref{Sec:source_and_target_data_prep}.

\emph{Reference source semantic segmentation network.}
Inspired by the semantic consistency loss from CyCADA ~\cite{hoffman2017cycada}, our design consists of two instances of SalsaNext. The first instance, which we call $f_{RefSrc}$, is decoupled from the rest of the system. We pre-train $f_{RefSrc}$ using supervised learning using the fully-labeled \emph{source} domain data set, and keep its parameters fixed after training. 

\emph{Target semantic segmentation network}.
We define $f_{target}$ as the semantic segmentation network that we wish to train to perform well on validation data from the target domain. We initialize $f_{target}$ with the weights from the pre-trained $f_{RefSrc}$, and split it into encoder $f_{enc}$ and decoder $f_{dec}$. Inspired by ~\cite{rochan2021unsupervised} and ~\cite{sun2020test}, we introduce an auxiliary decoder $Dec_{aux}$ (see Fig. \ref{HYLDA_architecture}). $Dec_{aux}$ is a SalsaNext decoder with a $tanh$ activation at the output, which we adapted to work on a simple auxiliary identity reconstruction task, which enables us to train $f_{enc}$ using self-supervision with \emph{all} of the available (labeled or unlabeled) target domain point clouds as explained below. We then train $f_{target}$ (both, $f_{enc}$ and $f_{dec}$ jointly) using both, semi-supervised training using the few available labeled frames, and unsupervised using fake translated images and labels from the \emph{source} domain. 

\section{Losses and Training Procedure} \label{Sec:training_and_losses}
In this section we describe the sequential steps ($6$ in total) that we execute for each batch of data (training step) during training. We also describe the losses employed at each of these training steps.
		
\emph{Step 1) Self-supervision of image-to-image translation generator encoders and decoders}.
This step is to regularize the mappings $G$ and $F$ as discussed above. First, we \emph{cross}-connect (concatenate) the skip connections (at two feature layer resolutions) between the source encoder $Enc_{s}$ and decoder $Dec_{s}$, and between the target encoder $Enc_{t}$ and decoder $Dec_{t}$ (see cross-connection arrows in Fig. \ref{HYLDA_architecture}). Then, we input the source image $y$ into $Enc_{s}$ so that $Dec_{s}$ generates a reconstructed identity image $y^*$. Similarly, a target image $x$ is fed into $Enc_{t}$ to obtain a reconstructed image $x^*$ at the output of $Dec_{t}$. We then use Eqn. (\ref{eq:gen_selfsuperv_loss}) to compute the self-supervision auxiliary identity $L1$ loss $\mathcal{L}_{xself}$ that we use to back-propagate through and update the encoders and decoders from both generators. This step explicitly encourages (using all available real data) the encoders to learn how to encode features from each source domain, while at the same time encouraging the decoders to learn how to generate reconstructed outputs, also from the source domain. This acts as a form of regularization since in the next training step the encoders and decoders will be trained using adversarial training without the cross-connections.
\begin{equation} 
\begin{aligned} 
\label{eq:gen_selfsuperv_loss}
\mathcal{L}_{xself} = & \EX_{y \sim p_{data}(y)} [ \| Dec_{s}(Enc_{s}(y)) - y \|_1 ] +  \\
                     & \EX_{x \sim p_{data}(x)} [ \| Dec_{t}(Enc_{t}(x)) - x \|_1 ].	
\end{aligned}
\end{equation}		
	
\emph{Step 2) Adversarial training of generators and discriminators }.
This step is the same adversarial training used by CycleGAN. Unlike in step $1$, here we now connect (concatenate) the skip connections between the source encoder $Enc_{s}$ and decoder $Dec_{t}$, and between the target encoder $Enc_{t}$ and decoder $Dec_{s}$. Then feed the source image $y$ through the source encoder $Enc_{s}$. The output from $Dec_{t}$ is a translated (fake) image $x^*$, which is fed into the discriminator $D_X$ to measure how realistic the fake image looks like compared to real images from the target domain. A similar process is repeated by feeding the target image $x$ through $Enc_{t}$ to obtain a translated (fake) image $y^*$ that is fed into the discriminator $D_Y$. We then use Eqn. (\ref{eq:LS_loss_G_and_F}) to compute the LSGAN losses:
\begin{equation}
\begin{aligned} 
\label{eq:LS_loss_G_and_F}
& \mathcal{L}_{LSGAN}(G,D_Y,X) =  \EX_{x \sim p_{data}(x)} [ (D_Y(G(x)) - 1)^2 ], \\
& \mathcal{L}_{LSGAN}(F,D_X,Y) =  \EX_{y \sim p_{data}(y)} [ (D_X(F(y)) - 1)^2 ],
\end{aligned}
\end{equation}

where $\mathcal{L}_{LSGAN}(G,D_Y,X)$ and $\mathcal{L}_{LSGAN}(F,D_X,Y)$ are the least squares GAN losses used to train the generators and discriminators as done in  ~\cite{mao2017least}, ~\cite{zhu2017unpaired}, ~\cite{corral-soto2021lcp}.
	
\emph{Step 3) Batch statistics loss for fake translated images}.	
As discussed in section \ref{Sec:image_to_image_engine}, to encourage the generators to output fake translated images that adhere to the underlying statistics from the \emph{real} input domain they are translated into, we compute the mean and covariance matrix from the fake translated image batches, and measure the statistics similarity loss $\mathcal{L}_{stats}$ with respect to the pre-computed training targets described in section \ref{Sec:image_to_image_engine} using Eqn.(\ref{eq:stats_loss}).	
\begin{equation}
\begin{aligned} 
\label{eq:stats_loss}
\mathcal{L}_{stats} = & ( \EX_{y \sim p_{data}(y)} [ \| \hat{\Sigma}_{s} - \Sigma_{s} \|_1 ]    +   \EX_{x \sim p_{data}(x)} [ \| \hat{\Sigma}_{t} - \Sigma_{t} \|_1 ]  ) + \\
                      & ( \EX_{y \sim p_{data}(y)} [ \| \hat{\mu}_{s} - \mu_{s} \|_1 ]     +   \EX_{x \sim p_{data}(x)} [ \| \hat{\mu}_{t} - \mu_{t} \|_1 ]  ).       
\end{aligned}
\end{equation}	
	
Finally, we compute the image-to-image loss $\mathcal{L}_{i2i}$ by adding the LSGAN and statistics loss using a scalar weight $\beta$ used to control the contribution from $\mathcal{L}_{stats}$ as shown in Eqn. (\ref{eqn:Loss_img2img}). We then back-propagate through and update the whole image-to-image translation engine (generators and discriminators) using this loss. We set $\beta=0.1$ experimentally.
\begin{equation} 
\begin{aligned} 
\mathcal{L}_{i2i} = & \mathcal{L}_{LSGAN}(G,D_Y,X) +  \mathcal{L}_{LSGAN}(F,D_X,Y) +  \beta \times \mathcal{L}_{stats}. 
\end{aligned}
\label{eqn:Loss_img2img}
\end{equation}

\emph{Step 4) Self-supervised training of target semantic segmentation encoder}.
As discussed in section \ref{Sec:problem_definition}, only a small number of frames are labeled in the target data set, and the rest are not labeled. However, we wish to exploit \emph{all} of the available frames (without the use of labels) to train $f_{enc}$ to learn features from the whole target domain data set. First, we connect all skip connections from $f_{enc}$ to the auxiliary identity reconstruction decoder $Dec_{aux}$. Then, we feed a \emph{real} target frame $x$ into $f_{enc}$, which propagates feature information into $Dec_{aux}$ to output a reconstructed $\tilde{x}$ (see Fig. \ref{HYLDA_architecture}). Next, we measure the identity reconstruction similarity $L1$ loss using Eqn.(\ref{eq:ssnet_selfsuperv_loss}), and back-propagate through and update both, $f_{enc}$ and $Dec_{aux}$. Note that we discard $Dec_{aux}$ after the last training epoch.   
\begin{equation}
\begin{aligned} 
\label{eq:ssnet_selfsuperv_loss}
\mathcal{L}_{ss-self} = & \EX_{x \sim p_{data}(x)} [ \| Dec_{aux}(f_{enc}(x)) - x \|_1 ].	
\end{aligned}
\end{equation}	

\emph{Step 5) Semi-supervised training of target semantic segmentation encoder and decoder}. 
In our problem, a small number (e.g. $100$) of target domain frames are labeled. We exploit these few labeled frames to update $f_{target}$ (both, $f_{enc}$ and $f_{dec}$) via standard supervised learning by computing the cross-entropy loss $\mathcal{L}_{wce}(y,\hat{y}_i)$ from Eqn.(\ref{eqn:salsanext_superv_crossentropy_loss}) as done in ~\cite{cortinhal2020salsanext}. 
\begin{equation}
\begin{aligned} 
\label{eqn:salsanext_superv_crossentropy_loss}
\mathcal{L}_{wce}(c,\hat{c}) = -\Sigma_{i} \alpha_i p(c_i) \log ( p(\hat{c}_i) )  \quad with \quad \alpha_i =1/\sqrt{f_i},
\end{aligned}
\end{equation}		
where $c_i$ and $\hat{c}_i$ define the true and predicted class labels, and $f_i$ is the frequency (number of points of the $i^{th}$ class), as defined by ~\cite{cortinhal2020salsanext}.
	
\emph{Step 6) Unsupervised training of target semantic segmentation encoder and decoder.}
Up to this point, $Enc_{s}$ and $Dec_{t}$ from the generator $F_{Y \to X}$, and $f_{target}$ are well-initialized, and partially trained by the training steps $1$ to $5$ explained above. Ideally, we would like to train $f_{target}$ with a larger labeled target domain data set, however we don't have access to such large labeled data set. To mitigate this limitation, our strategy is to train $f_{target}$ with the whole labeled training set from the \emph{source} domain, but translated by the generator $F_{Y \to X}$ into ~\emph{fake} target domain frames, while leveraging the labels from the source domain.

\emph{Leveraging the labels from the source domain.} The spatial structure consistency between the fake image $x^*$ and the source domain label map is encouraged in two ways: 1) By the skip connections between $Enc_{s}$ and $Dec_{t}$, and between $Enc_{t}$ and $Dec_{s}$, which propagate spatial information into the decoders, and 2) Through the semantic task conistency loss $\mathcal{L}_{sem}$, Eqn. (\ref{eq:semantic_consistency_loss}), which encourages $f_{target}$ to output semantic maps that are spatially consistent (structure) with the semantic maps from $f_{RefSrc}$. This enables us to leverage the source labels and use them with the fake translated images $x^*$.
On the other hand, we wish to use $f_{target}$ to help supervise the training of the image-to-image translation engine, specifically $Enc_{s}$ and $Dec_{t}$. For this, we input a real source image $y$ into the pre-trained $f_{RefSrc}$. Then, input a translated \emph{fake} target image $x^*$ into the network $f_{target}$. Both of them predict a semantic segmentation map as their output. We use these maps to compute two losses  (see Fig. \ref{HYLDA_architecture}):  1) The segmentation task cross-entropy loss (Eqn.(\ref{eq:salsanext_unsuperv_crossentropy_loss})) with respect to the \emph{source} ground-truth label maps, and 2) The semantic task consistency loss $\mathcal{L}_{sem}$ which encourages $f_{target}$ to output segmentation maps that are as similar as possible to those from $f_{RefSrc}$. We aggregate these two losses using Eqn.(\ref{eq:usupervised_loss}), where $\gamma$ is a scalar to control the contribution from the semantic consistency term, which we set to $1$ experimentally. We use $\mathcal{L}_{unsupervised}$ to back-propagate through the image-to-image translation engine and update it so that it learns to generate better fake translated images in the next iteration. After this we also update $f_{target}$ using the  translated \emph{fake} target image batch $x^*$ to improve its generalization.
\begin{equation}
\begin{aligned} 
\label{eq:salsanext_unsuperv_crossentropy_loss}
\mathcal{L}_{uwce}(c,\hat{c}) = -\Sigma_{i} \alpha_i p(c_i) \log ( p(\hat{c}_i) ).
\end{aligned}
\end{equation}	
	
\begin{equation} 
\begin{aligned} 
\label{eq:semantic_consistency_loss}
\mathcal{L}_{sem} = & \EX_{y \sim p_{data}(y)} [ \| f_{RefSrc}(y) - f_{target}(x^*) \|_1 ].	
\end{aligned}
\end{equation}	
	
\begin{equation} 
\begin{aligned} 
\label{eq:usupervised_loss}
\mathcal{L}_{unsupervised} = \mathcal{L}_{uwce}(y,\hat{y}_i)  + \gamma \times \mathcal{L}_{sem}.
\end{aligned}
\end{equation}		

\subsection{Training Details} \label{Sec:training_details}
To train HYLDA, we use the Adam optimizer with a constant learning rate of $\eta_{ss} = 0.01$ for $f_{target}$, $f_{RefSrc}$ and $Dec_{aux}$ with data augmentation as done in ~\cite{cortinhal2020salsanext}. The image-to-image translation engine is trained using Stochastic Gradient Descent (SGD) with a constant learning rate of $\eta_{i2i} = 0.002$. We train HYLDA for \textcolor{black}{$75$} epochs on Nvidia Tesla V100 GPUs, each with 32510 MB memory. Training on 8 GPUs with batch size 2 per GPU, takes approx. 2 hours per epoch for the domain adaptation in the SemanticKITTI $\to$ nuScenes direction. 

\section{Experiments and Evaluations} \label{Sec:Experiments}
\begin{figure*} 
	\centering	
	\includegraphics[width=0.8\columnwidth, trim={2cm 15cm 2cm 0cm},clip]{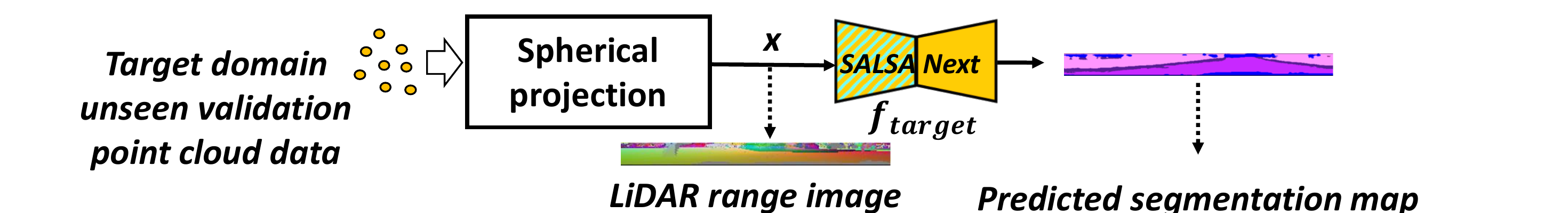} 		
	\caption{During inference, the range image projection of a point cloud from target domain is fed to our trained LiDAR segmentation network that outputs its semantic segmentation map  		     				
	\label{inference_mode} }
\end{figure*}

In inference mode, unseen frames $x$ from the target domain \emph{validation} data set are directly fed into $f_{target}$ as shown in Fig. \ref{inference_mode}. 

\subsection{Source and Target Domain Dataset Preparation} \label{Sec:source_and_target_data_prep}
LiDAR semantic segmentation networks such as ~\cite{cortinhal2020salsanext}, ~\cite{wu2019squeezesegv2}, ~\cite{gerdzhev2020tornado} are usually trained and evaluated on publicly-available data sets such as SemanticKITTI ~\cite{behley2019semantickitti} and nuScenes ~\cite{caesar2020nuscenes}. We choose the same data sets to develop and evaluate our HYLDA domain adaptation method. SemanticKITTI has $19130$ labeled frames (point-level semantic class ID) for training, and $4071$ labeled validation frames for evaluation. nuScenes has $28130$ labeled frames for training, and $6019$ labeled validation frames. Since these data sets have different class definitions, we follow prior work ~\cite{rochan2021unsupervised} to define the following $11$ overlapping object classes in these data sets: $\{$ Car, Bicycle, Motorcycle, Other vehicle, Pedestrian, Truck, Drivable surface, Sidewalk, Terrain, Vegetation, Man-made $\}$. All other objects fall under a background class. To train HYLDA, we utilize all of the available frames and labels from both $Y$ (source) and $X$ (target) data sets. Let $N=|Y|$ and $M=|X|$ be the number of available raw data frames for the source and target domains respectively. And let $N_{gt}$ and $M_{gt}$ be their corresponding number of labeled frames. For the source domain $Y$ all frames are labeled ($N=N_{gt}$). In our problem, we are given only a few labeled target domain frames ($M_{gt} \ll M$). To simulate the availability of a small number of labeled frames using SemanticKITTI and nuScenes (where all training frames are labeled), we use random sampling (uniform distribution) of the target set $X$ to define the training subsets  $X_{100} \in X_{250} \in X_{500}$, where the subscripts denote the number of labeled frames.

\subsection{Baseline Preparation} \label{Sec:baseline_prep}
We compared our HYLDA method against two baselines. The first baseline is SalsaNext fine-tuning, where we initialize the network $f_{target}$ with the pre-trained source $f_{RefSrc}$ network weights, and train/refine $f_{target}$ with each available subset of labeled target frames ($X_{100}$, $X_{250}$, and $X_{500}$) for $100$ epochs.
For the second baseline we adopted the recent method from ~\cite{corral-soto2021lcp} called LCP, which uses CycleGAN ~\cite{zhu2017unpaired} for image-to-image translation and works with the PointPillars 3D object detector ~\cite{lang2019pointpillars}. We re-implemented the LCP and adapted it to work with
$64$ $\times$ $2048$ $\times 5$ LiDAR range view images for SalsaNext ~\cite{cortinhal2020salsanext} semantic segmentation. Note that the LCP baseline does not use any of the auxiliary decoder $Dec_{aux}$, reference network $f_{RefSrc}$, or semantic consistency loss. Moreover, its CycleGAN generators don't have skip connections, and the discriminators are a single instance of the Patch-GAN discriminator working at a single resolution.

\subsection{Evaluation} \label{Sec:evaluation}
\begin{figure*} 
	\centering	
	\includegraphics[width=0.8\columnwidth, trim={3cm 20.5cm 3cm 0cm},clip]{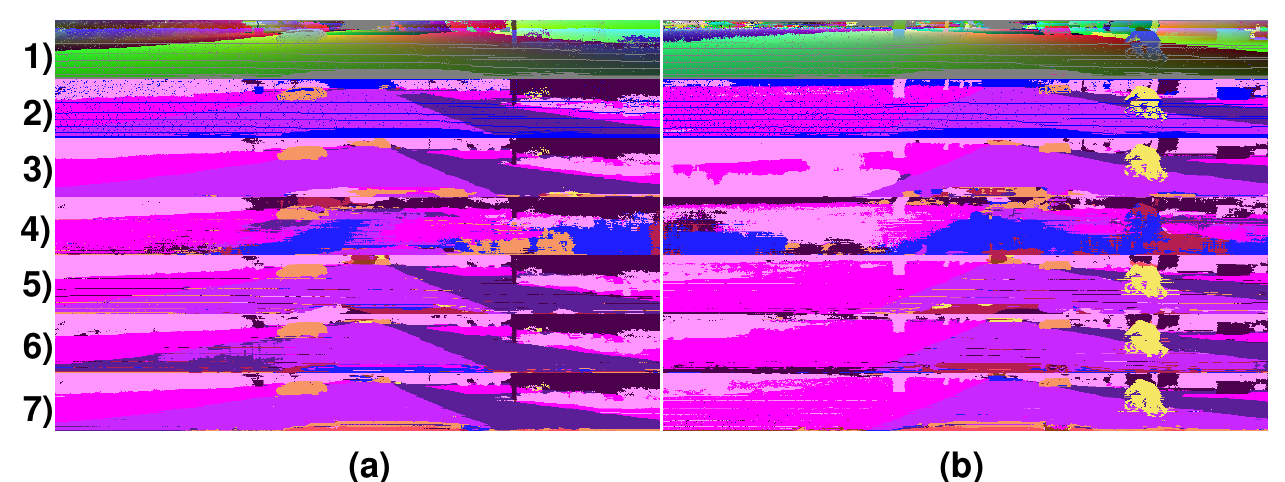} 		
	\caption{Example qualitative outputs for nuScenes $\to$ SemanticKITTI $X_{500}$.  Rows: 1) Target domain validation input, 2) Ground-truth, 3) SalsaNext Oracle, 4) SalsaNext Na\"ive, 5) SalsaNext Fine-tuning, 6) LCP ~\cite{corral-soto2021lcp}, 7) HYLDA  		     				
	\label{Example_ouput_segm_maps} }
\end{figure*}

We trained HYLDA in two domain adaptation directions: SemanticKITTI $\to$ nuScenes (K $\to$ N), and nuScenes $\to$ SemanticKITTI (N $\to$ K), with the three small labeled target domain subsets $X_{100}$, $X_{250}$, and $X_{500}$. We evaluate on the full validation data set from the target domain using the Mean Intersection over the Union (mIoU) metric as done in ~\cite{cortinhal2020salsanext}, ~\cite{wu2019squeezesegv2}, ~\cite{gerdzhev2020tornado}. The results are summarized in Table \ref{table:kitti_vs_nuscenes_eval}.
\textcolor{black}{Rows 1 and 14} show the results from the SalsaNext oracles (we follow the methodology from ~\cite{hoffman2017cycada} and ~\cite{zhu2017unpaired}), where $f_{target}$ is trained with all of the \emph{target} $M=M_{gt}$ labeled frames and evaluated on the target validation data set. \textcolor{black}{Rows 2 and 15} show the results of SalsaNext na\"ive training with the $N=N_{gt}$ source labeled frames and evaluating on the target validation data set. Note that training on the source domain and directly evaluating on the target domain results in poor generalization, which motivates the study of domain adaptation methods. \textcolor{black}{Rows 3,4,16,17} show results from two unsupervised methods,  LiDAR UDA ~\cite{rochan2021unsupervised} and DeepCORAL ~\cite{sun2016deep} included for reference. \textcolor{black}{Rows 5 through 13, and 18 through 26} show the results obtained with the SalsaNext fine-tuning and LCP baselines, as well as with our HYLDA method. We observe that none of the models was able to reach the oracle performance, which is expected. We further observe that the performance of all models tends to improve as the number of labeled target frames is increased. For the K $\to$ N direction, both, the LCP and HYLDA are consistently better than SalsaNext fine-tuning. On average, HYLDA outperforms SalsaNext fine-tuning by $8.3\%$ in the $K \to N$ direction, and by $5.3\%$ in the $N \to K$ direction. For the $K \to N$ direction HYLDA outperforms the LCP on average by $2.36\%$. On the other hand, for the nuScenes $\to$ SemanticKITTI direction we observe that the LCP results are inferior to the SalsaNext fine-tuning baseline. By looking at Table \ref{table:kitti_vs_nuscenes_eval}, we observe that LCP had problems with four classes: bicycles, motorcycles, trucks, and other vehicles, whereas HYLDA achieves a much stronger performance, outperforming LCP by an average of $12.3\%$, demonstrating the importance of its modules.  Fig. \ref{Example_ouput_segm_maps} shows qualitative results for two frames from the target validation set. In this figure, we observe that the segmentation maps produced by $f_{target}$ trained within HYLDA are more accurate and visually closer to the oracle than the alternative methods.  

\begin{table*}[h!]
	\caption{Domain adaptation results for both SemanticKITTI $\to$ nuScenes ($K \to N$), and nuScenes $\to$ SemanticKITTI ($N \to K$) directions, with $100$, $250$, and $500$ labeled target frames for semi-supervised training. We evaluate on the target validation data set using the Mean Intersection over the Union (mIoU) as done in ~\cite{cortinhal2020salsanext}, ~\cite{wu2019squeezesegv2}, ~\cite{gerdzhev2020tornado}. * We re-implemented the LCP method from ~\cite{corral-soto2021lcp} to work with SalsaNext semantic segmentation }
	\label{table:kitti_vs_nuscenes_eval}
	\begin{threeparttable}
		\centering
		\renewcommand{\arraystretch}{1.3}
		\adjustbox{max width=\textwidth}{%
			\begin{tabular}{|lc|c|ccccccccccc|c|c|}
				\hline
				\rotatebox{0}{Model/experiment} & 
				\rotatebox{0}{Direction} & 
				\rotatebox{90}{\# trg frames} &
				\rotatebox{90}{Car} &
				\rotatebox{90}{Bicycle} &
				\rotatebox{90}{Motorcycle} &
				\rotatebox{90}{Other veh} &
				\rotatebox{90}{Pedestrian} &
				\rotatebox{90}{Truck} &
				\rotatebox{90}{Driv. Surf} &
				\rotatebox{90}{Sidewalk} &
				\rotatebox{90}{Terrain} &
				\rotatebox{90}{Vegetation} &
				\rotatebox{90}{Manmade} &
				\multicolumn{1}{|c|}{\rotatebox{0}{mIoU}} \\ \hline
				1) SalsaNext~\cite{cortinhal2020salsanext} Oracle    & N $\to$ N  & 28K   & 84.5 & 22.7 & 66.4 & 63.3 & 59.5 & 72.7 & 96.4 & 73.4 & 74.0 & 85.4 & 87.9 & 71.5 \\ \hdashline
				2) SalsaNext Na\"ive        & K $\to$ N  &  0    & 42.8 & 0.17 & 0.09 & 4.69 & 11.6 & 3.75 & 50.3 & 17.0 & 26.2 & 15.3 & 43.5 & 19.6 \\ \hdashline
                     3) LiDAR UDA ~\cite{rochan2021unsupervised} & K $\to$ N     & 28K       & 54.4 & 3.0 & 1.9 & 7.6 & 27.7 & 15.8 & 82.2 & 29.6 & 34.0 & 57.9 & 65.7 & 34.5 \\
                     4) DeepCORAL ~\cite{sun2016deep}  &     &       & 51.0 & 0.9 & 6.0 & 4.0 & 25.9 & 29.9 & 82.6 & 27.1 & 27.0 & 55.3 & 56.7 & 33.3 \\ \hdashline
				5) SalsaNext Fine-tuning  & K $\to$ N  & 100   & 65.9 & 1.83 & 0.57 & 11.1 & 19.3 & 22.4 & 69.6 & 41,7 & 47.0 & 64.4 & 60.7 & 36.8 \\ 
				6) LCP* ~\cite{corral-soto2021lcp}   &                          &       & 69.2 &	7.77 & 10.9 & 28.5 & 47.0 &	34.0 & 71.9 & 45.2 & 59.8 & 71.0 & 65.0 & 46.0 \\ 
				7) HYLDA                       &                          &       & 69.0 & 5.52 & 9.39 & 23.5 & 45.4 & 37.7 & 78.8 & 56.1 & 61.8 & 74.7 & 69.0 & \textbf{48.3} \\ \hdashline				
				8) SalsaNext Fine-tuning  & K $\to$ N           & 250   & 70.7 & 2.79 & 9.43 & 19.0 & 28.2 & 32.5 & 73.8 & 49.3 & 54.7 & 71.9 & 65.6 & 43.4 \\ 
				9) LCP   &                          &       & 74.0 &	3.23 & 14.4 & 27.7 & 45.8 &	35.3 & 76.2 & 51.5 & 59.2 &	70.5 & 66.7 & 47.6 \\ 
				10) HYLDA                       &                          &       & 71.6 & 9.00 & 22.2 & 28.7 & 48.8 & 35.1 & 79.3 & 57.3 & 62.8 & 72.1 & 66.8 & \textbf{50.3} \\ \hdashline				
				11) SalsaNext Fine-tuning  & K $\to$ N           & 500   & 73.0 & 2.41 & 10.7 & 24.0 & 35.5 & 14.0 & 78.5 & 54.8 & 59.5 & 75.6 & 70.0 & 45.8 \\ 
				12) LCP  &                          &       & 73.7 &	5.43 & 9.33 & 30.0 & 47.2 &	36.9 & 80.8 & 59.8 & 65.2 &	73.9 & 70.0 & 50.2 \\ 
				13) HYLDA                     &                          &       & 67.8 & 7.4  & 36.6 & 26.0 & 49.2 & 34.3 & 82.0 & 59.0 & 61.5 & 78.0 & 73.4 & \textbf{52.3} \\  				
				\hline 
				\hline
				14) SalsaNext~\cite{cortinhal2020salsanext} Oracle         & K $\to$ K  & 19K   & 92.2 & 52.6 & 47.8 & 48.3 & 53.7 & 80.2 & 94.6 & 82.5 & 70.6 & 85.9 & 86.8 & 72.3 \\ \hdashline
				15) SalsaNext Na\"ive       & N $\to$ K  &  0    & 6.53 & 0.94 & 0.55 & 0.77 & 6.61 & 0.26 & 20.5 & 3.23 & 22.1 & 23.9 & 22.0 & 9.77 \\ \hdashline
                     16) LiDAR UDA ~\cite{rochan2021unsupervised} & K $\to$ N     & 19K    & 49.6 & 4.6 & 6.3 & 2.0 & 12.5 & 1.8 & 25.2 & 25.2 & 42.3 & 43.4 & 45.3 & 23.5 \\
                     17) DeepCORAL ~\cite{sun2016deep}  &     &       & 47.3 & 10.4 & 6.9 & 5.1 & 10.8 & 0.7 & 24.8 & 13.8 & 31.7 & 58.8 & 45.5 & 23.2 \\ \hdashline
				18) SalsaNext Fine-tuning & N $\to$ K  & 100   & 82.7 & 18.4 & 3.81 & 5.47 & 22.4 & 10.4 & 69.9 & 54.0 & 49.8 & 43.4 & 56.4 & 37.9 \\ 
				19) LCP* ~\cite{corral-soto2021lcp}  &                          &       & 70.2 &	3.45 & 12.3	& 1.67 & 12.9 &	6.94 & 66.2 & 49.3 & 37.5 & 34.2 & 48.0 & 31.2 \\ 
				20) HYLDA                      &                          &       & 87.3 & 29.2 & 26.9 & 24.4 & 34.6 &	27.6 & 70.3 & 52.3 & 47.4 &	46.4 & 61.7 & \textbf{46.2} \\ \hdashline				
				21) SalsaNext Fine-tuning & N $\to$ K            & 250   & 82.2 & 37.0 & 5.94 & 18.3 & 32.4 & 20.2 & 74.8 & 56.7 & 53.8 & 50.1 & 60.6 & 44.7 \\ 
				22) LCP   &                         &       & 72.6 & 16.6 & 15.7 & 18.8 & 26.6 &	16.2 & 66.6 & 52.1 & 51.0 &	46.9 & 57.7 & 40.0 \\ 
				23) HYLDA                      &                          &       & 88.2 &	29.8 & 26.0 & 31.4 & 43.6 &	24.8 & 73.1 & 57.0 & 52.4 & 51.0 & 63.5 & \textbf{49.2} \\ \hdashline				
				24) SalsaNext Fine-tuning & N $\to$ K            & 500   & 88.8 & 39.3 & 20.8 & 18.0 & 42.0 & 39.5 & 75.0 & 57.8 & 54.0 & 52.0 & 66.0 & 50.3 \\ 
				25) LCP   &                          &       & 78.5 & 28.6 & 20.2 & 20.4 & 30.3 & 6.06 & 65.3 & 49.5 & 47.6 & 45.8 & 57.9 & 40.9 \\ 
				26) HYLDA                       &                          &       & 89.1 &	41.1 & 35.7 & 36.2 & 47.0 & 26.0 & 79.8 & 63.9 & 52.0 &	49.9 & 68.9 & \textbf{53.6} \\  				
				\hline
			\end{tabular}%
		}
	\end{threeparttable}
\end{table*}

\emph{Validating HYLDA on a Third Data Set.}
To further validate our method, we perform domain adaptation experiments using a third LiDAR data set for autonomous driving: SemanticPoss ~\cite{pan2020semanticposs}.  SemanticPoss was captured at the Peking University in China. It has a total of $2988$ LiDAR outdoor scans captured using a 40-channel LiDAR sensor. It comes with $14$ labeled classes, and the main strength ~\cite{pan2020semanticposs} of the data set, compared to other data sets such as SemanticKITTI and nuScenes, is a higher average number of instances per frame for the people and rider classes. We partition the data set into $1988$ training and $1000$ validation frames ~\cite{alonso2020domain}. From the training set, we build a target domain training subset $X_{100}$, by random sampling of $100$ frames. We perform experiments in the SemanticKITTI $\to$ SemanticPoss and SemanticPoss $\to$ SemanticKITTI directions. The results are shown in Table \ref{table:kitti_vs_poss_eval}. Once again, we observe that our HYLDA method outperforms the baselines. In the SemanticKITTI $\to$ SemanticPoss direction it outperformed Fine-tuning by $1\%$ and LCP by $1.8\%$. And in the other direction by $8.8\%$ and $12.5\%$ respectively. In this SemanticPoss $\to$ SemanticKITTI direction we notice that the performance was greatly improved for the Pedestrian, Rider, and Bike classes, which is conistent with the SemanticPoss data set strengths described above. Moreover, not many instances of these classes are available in SemanticKITTI.  

\emph{Statistics of fake images translated by HYLDA.}
Table \ref{table:kitti_vs_poss_eval} shows that the SemanticPoss source domain data was successfully translated by our HYLDA domain adaptation method into \emph{fake} SemanticKITTI target domain data that was used to train $f_{target}$. To further confirm and understand the domain adaptation process, we used our \emph{trained} generator $F_{Y \to X}$ to run experiments in inference mode. We translated the SemanticPoss validation data set into fake SemanticKITTI data (and vice versa). We then computed the Mean Absolute Error (MAE) between the covariance matrices and mean range images of the \emph{fake} target outputs against those from the \emph{real} target domain data.  The results are shown in Table \ref{table:kitti_vs_poss_covmtx_mae}. Rows $1$ and $3$ show the MAE for the reference na\"ive scenario of no domain adaptation. Rows $2$ and $4$ show the effects after training HYLDA (including $F_{Y \to X}$) on the statistics of the fake translated images. We observe how the statistics of the \emph{fake} translated data become close to those from the \emph{real} target domain after training, especially for the case of the covariance matrix ($0.4317 \to 0.0650$ between rows $1$ and $2$, and $0.5244 \to 0.0412$ between rows $3$ and $4$  ). For the mean image MAE, the reduction of $0.0265 \to 0.0228$ between rows $1$ and $2$ is smaller than that of $0.0281 \to 0.0073$ between rows $3$ and $4$. We believe that this relates to the smaller size of the SemanticPoss training data set ($1.9K$ frames), compared to $19K$ frames from SemanticKITTI. Nevertheless, these numbers confirm the effectiveness of the domain adaptation performed by HYLDA. The effects of different data set sizes and number of class instances needs to be investigated in future work.

\begin{table*}[h!]
	\caption{Domain adaptation results for both SemanticKITTI $\to$ SemanticPoss ~\cite{pan2020semanticposs} ($K \to P$), and SemanticPoss $\to$ SemanticKITTI ($P \to K$) directions, with $100$ labeled target frames ($X_{100}$) for semi-supervised training}
	\label{table:kitti_vs_poss_eval}
	\begin{threeparttable}
		\centering
		\renewcommand{\arraystretch}{1.3}
		\adjustbox{max width=\textwidth}{%
			\begin{tabular}{|lc|c|ccccccccccc|c|c|}
				\hline
				\rotatebox{0}{Model/experiment} & 
				\rotatebox{0}{Direction} & 
				\rotatebox{90}{\# trg frames} &
				\rotatebox{90}{Pedestrian} &
				\rotatebox{90}{Rider} &
				\rotatebox{90}{Car} &
				\rotatebox{90}{Trunk} &
				\rotatebox{90}{Plants} &
				\rotatebox{90}{Traffic sign} &
				\rotatebox{90}{Pole} &
				\rotatebox{90}{Building} &
				\rotatebox{90}{Fence} &
				\rotatebox{90}{Bike} &
				\rotatebox{90}{Ground} &
				\multicolumn{1}{|c|}{\rotatebox{0}{mIoU}} \\ \hline
				1) SalsaNext~\cite{cortinhal2020salsanext} Oracle         & P $\to$ P  & 1.9K   & 58.8 & 42.1 &	90.0 &	21.4 &	74.5	 &  48.4  &42.9   & 83.8  & 49.8   &  67.9 & 72.8   &	59.3 \\ \hdashline
				2) SalsaNext Na\"ive        & K $\to$ P  &  0     & 34.1 &	11.8 &  47.5 &  10.8 &	28.4 & 8.18   &14.2    & 56.8  &  3.57   &  21.4 & 72.8   &	28.1 \\ \hdashline
				3) SalsaNext Fine-tuning   & K $\to$ P  & 100   & 57.3 &	40.9 &	88.9 &	20    &	73.6 & 43.2  & 35.6    &	84.1 &	39.0 &	64.7 &	72.2 &	56.3 \\ 
				4) LCP ~\cite{corral-soto2021lcp}  &                 &         & 63.0  &	27.1 &	80.7 &	30    &	 72.3 &	28.3 &	33.3 &	80.1 &	48.7  &	70.5 &	77.0	  &  55.5 \\ 
				5) HYLDA                        &                 &         & 60.4 &	44.4 &	86.6 &	24.6 &	74.2  &	38.0 &	26.4 &	84.7 &	42.3 &	71.5 &	76.8 &	\textbf{57.3} \\ \hdashline				
				\hline 
				\hline
				6) SalsaNext~\cite{cortinhal2020salsanext} Oracle         & K $\to$ K  & 19K   & 62.7 & 68.6  &	93.9 &	64.4 &	56.7 &	45.8 &	59.9 &	77.9 &	53.5 &	53.5 &	71.8 &	64.4 \\ \hdashline
				7) SalsaNext Na\"ive       & P $\to$ K  &  0      & 2.00  &	0        &	47.3  &	1.80  &	27.7 &	4.85  &	12.1 &	25.2 &	0     &	0.13  &	6.98  &	11.6 \\ \hdashline
				8) SalsaNext Fine-tuning  & P $\to$ K &  100   & 29.9 &	34.7  &	85.9  &	49.7 &	39.6 &	25.7 &	48.8 &	59.5 &	28.0 &	0.01  &	58.4 &	41.8 \\ 
				9) LCP ~\cite{corral-soto2021lcp}   &               &         & 27.6 &	13.7  &	85.3 &	45.8 &	34.9 &	33.6 &	54.5 &	58.4 &	15.3 &	1.16  &	48.7 &	38.1 \\ 
				10) HYLDA                      &               &         & 42.6 &	48.3  &	91.6 &	55.5 &	48.8 &	30.9 &	55.1  &	68.6  &	36.0 &	13.5 &	65.8 & \textbf{50.6} \\ \hdashline				
				\hline
			\end{tabular}%
		}
	\end{threeparttable}
\end{table*}

\setlength{\tabcolsep}{4pt}
\begin{table}
\begin{center}
\caption{HYLDA domain adaptation effects on the statistics of the translated data. We compute the mean absolute error (MAE) between the covariance matrices and mean range images of the \emph{fake translated images} outputted by the generator $F_{Y \to X}$ agaist those from the \emph{real} target domain (see rows $2$ and $4$). We compare against the Na\"ive (no-translation case) to see the effect before and after the training process}
\label{table:kitti_vs_poss_covmtx_mae}
\begin{tabular}{llll}
\hline\noalign{\smallskip}
Data to compare                             & Target domain               & Mean                  &   Cov.Matrix\\
against the target domain                &      (Real)                      &  MAE                  &   MAE\\
\noalign{\smallskip}
\hline
\noalign{\smallskip}
1) Sem. Poss val set (real)              & Sem. KITTI train set       & 0.0265                & 0.4317 \\
2) Fake Sem. KITTI                        & Sem. KITTI train set       & \textbf{0.0228}   & \textbf{0.0650}\\
\hline
3) Sem. KITTI val set (real)            & Sem. Poss train set         &0.0281                 & 0.5244 \\
4) Fake Sem. Poss                          & Sem. Poss train set         & \textbf{0.0073}   & \textbf{0.0412}\\
\hline
\end{tabular}
\end{center}
\end{table}
\setlength{\tabcolsep}{1.3pt}

\subsection{Ablation Study} \label{Sec:ablation_study}
\begin{wrapfigure}{r}{0.5\textwidth}
  \centering
     \includegraphics[width=0.5\columnwidth, trim={0 170 0 170},clip]{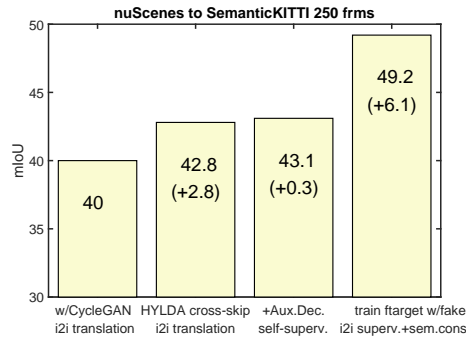}
	\caption{Ablation study to determine the contribution from key modules in HYLDA. We conduct this study from nuScenes $\to$ SemanticKITTI direction, for the case when we have access to $250$ labeled frames from SemanticKITTI}
	\label{fig:eval_bar_plots_ablation} 	
\end{wrapfigure}

We conduct an ablation study on the N $\to$ K direction with the $X_{250}$ labeled target domain semi-supervision data set. We start with using the LCP CycleGAN-based image-to-image translation (instead of ours). Then, we enable the following elements of HYLDA incrementally:  1) Use our new HYLDA image-to-image translation engine (instead of CycleGAN), including skip connection mechanism and statistics loss, 2) Auxiliary decoder (self-supervision of $f_{enc}$), and 3) Unsupervised training of $f_{target}$ with fake images from the generator $F_{Y \to X}$, including semantic consistency loss, and supervision of image-to-image translation by $f_{target}$. 
The bar plot in Fig. \ref{fig:eval_bar_plots_ablation} summarizes the results of this study. We observe that replacing the CycleGAN image-to-image translation with our proposed image-to-image translation engine results in a clear improvement of $2.8\%$. Enabling self-supervision ($f_{end}$ and $Dec_{aux}$) results in a modest contribution of $0.3\%$. However, when the unsupervised training of $f_{target}$ with fake data and supervision of image-to-image translation engine is enabled (i.e. full HYLDA system), the performance improved by $6.1\%$, demonstrating the importance of these elements of HYLDA. On the flip side, we did not observe significant performance contribution from the dual-head discriminator when evaluated independently on LiDAR range images. We believe that its contribution might become more noticeable in domain adaptation for computer vision with camera images where the data is dense. 

\section{Conclusions} \label{Sec:conclusions}
We have presented HYLDA, a novel end-to-end integrated domain adaptation framework for semantic segmentation of LiDAR range images that successfully addressed the challenging problem of improving generalization on validation data from the target domain with only a few available target domain labeled frames. We have demonstrated the effectiveness of our method on three publicly-available LiDAR data sets, where HYLDA outperformed two domain adaptation baselines. We have also conducted ablation studies which showed that the main performance contributions in HYLDA can be attributed to our new image-to-image translation engine and to the use of the unsupervised training of the task semantic segmentation network with fake translated data, and to its role in the supervision of the image-to-image translation engine. We believe that our framework and training strategy are generic enough to be used in other applications and tasks, such as 3D object detection and panoptic segmentation, as well as in computer vision tasks.

%

\clearpage
%
%
\bibliographystyle{splncs04}
\bibliography{hylda_master_egbib_references}
\end{document}